\pdfoutput=1
\documentclass[10pt,twocolumn,letterpaper]{article}

\usepackage{cvpr}
\usepackage{times}
\usepackage{epsfig}
\usepackage[numbers,sort,compress]{natbib}
\usepackage{graphicx}
\usepackage{amsmath}
\usepackage{amssymb}
\usepackage{amsthm}

\usepackage[pagebackref=true,breaklinks=true,letterpaper=true,colorlinks,bookmarks=false]{hyperref}

\cvprfinalcopy % *** Uncomment this line for the final submission

\def\cvprPaperID{****} % *** Enter the CVPR Paper ID here

\newcommand{\cS}{\mathcal{S}}

\begin{document}

%%%%%%%%% TITLE
\title{Any-Shot Sequential Anomaly Detection in Surveillance Videos}

\author{Keval Doshi\\
University of South Florida\\
4202 E Fowler Ave, Tampa, FL 33620\\
{\tt\small kevaldoshi@mail.usf.edu}
\and
Yasin Yilmaz\\
University of South Florida\\
4202 E Fowler Ave, Tampa, FL 33620\\
{\tt\small yasiny@usf.edu}
}

\maketitle
%\thispagestyle{empty}

%%%%%%%%% ABSTRACT
\begin{abstract}
Anomaly detection in surveillance videos has been recently gaining attention. Even though the performance of state-of-the-art methods on publicly available data sets has been competitive, they demand a massive amount of training data. Also, they lack a concrete approach for continuously updating the trained model once new data is available. Furthermore, online decision making is an important but mostly neglected factor in this domain. Motivated by these research gaps, we propose an online anomaly detection method for surveillance videos using transfer learning and any-shot learning, which in turn significantly reduces the training complexity and provides a mechanism which can detect anomalies using only a few labeled nominal examples. Our proposed algorithm leverages the feature extraction power of neural network-based models for transfer learning, and the any-shot learning capability of statistical detection methods.
\end{abstract}

%%%%%%%%% BODY TEXT

\vspace{-5mm}
\section{Introduction}

The rapid advancements in the technology of closed-circuit television (CCTV) cameras and their underlying infrastructure has led to a sheer number of surveillance cameras being implemented globally, estimated to go beyond 1 billion by the end of the year 2021 \cite{videosurveillance}. Considering the massive amounts of videos generated in real-time, manual video analysis by human operator becomes inefficient, expensive, and nearly impossible, which in turn makes a great demand for automated and intelligent methods for an efficient video surveillance system. An important task in video surveillance is anomaly detection, which refers to the identification of events that do not conform to the expected behavior \cite{chandola2009anomaly}. %However, supervised classification algorithms fail to tackle the problem as it is almost impossible to define an anomalous event which takes all possible anomalous patterns/behaviors into account \cite{sultani2018real}.

A vast majority of the recent video anomaly detection methods directly depend on deep neural network architectures \cite{sultani2018real}. It is well known that these deep neural network models are data hungry. As a result, they require many labeled nominal frames and long hours of training to produce acceptable results on a new data set. Moreover, most of them are not suitable for online detection of anomalies as they need the knowledge of future video frames for appropriate normalization of detection score. 

Motivated by the aforementioned domain challenges and research gaps, we propose a hybrid use of neural networks and statistical $k$ nearest neighbor ($k$NN) decision approach for finding video anomalies with limited training in an online fashion. In summary, our contributions in this paper are as follows:
\begin{itemize}
    \item We significantly reduce the training complexity by leveraging \emph{transfer learning} while simultaneously outperforming the current state-of-the-art algorithms.
    \item We propose a novel framework for statistical \emph{any-shot} sequential anomaly detection which is capable of learning \emph{continuously} and from \emph{few samples}.
    \item Extensive evaluation on publicly available data sets show that our proposed framework can transition effectively between \emph{few-shot} and \emph{many-shot} learning.
\end{itemize}
\section{Related Works}
\label{s:Related}
Anomaly detection methods for video surveillance can be broadly classified into two categories: traditional and deep learning-based methods. Traditional methods  \cite{saligrama2012video,dalal2005histograms,zhao2011online,mo2013adaptive} extract hand-crafted motion and appearance features such as histogram of optical flow \cite{chaudhry2009histograms,colque2016histograms} and histogram of oriented gradients \cite{dalal2005histograms} to detect spatiotemporal anomalies \cite{saligrama2012video}. The recent literature is dominated by the neural network-based methods \cite{hasan2016learning,hinami2017joint,liu2018future,luo2017revisit,ravanbakhsh2019training,sabokrou2018adversarially,xu2015learning} due to their superior performance \cite{xu2015learning}. For instance, in \cite{luo2017remembering}, Convolutional Neural Networks (CNN), and Convolutional Long Short Term Memory (CLSTM) are used to learn appearance and motion features. More recently, Generative Adversarial Networks (GANs) have been used to generate internal scene representations based on a given frame and its optical flow to detect deviation of the GAN output from the nominal data \cite{ravanbakhsh2017abnormal,liu2018future}. However, there is also a significant debate on the shortcomings of neural network-based methods in terms of interpretability, analyzability, and reliability of their decisions \cite{jiang2018trust}. For example, \cite{papernot2018deep,sitawarin2019defending} propose using a nearest neighbor-based approach together with deep neural network structures to achieve robustness, interpretability for the decisions made by the model, and defense against adversarial attack. Also, deep neural networks for visual recognition typically  require  a  large  amount  of  labelled  examples for training \cite{krizhevsky2012imagenet}, which might not be available for all possible behaviors/patterns. Hence, recently researchers have begun to address the challenge of few-shot learning \cite{koch2015siamese,sung2018learning,snell2017prototypical,vinyals2016matching}. A line of few-shot learning methods is based on the idea of transfer learning, i.e, using a pre-trained model learned from one domain for another domain \cite{pan2009survey,yosinski2014transferable,kornblith2019better}.    
\section{Proposed Method}

\begin{figure*}[th]
\centering
\includegraphics[width=1\textwidth]{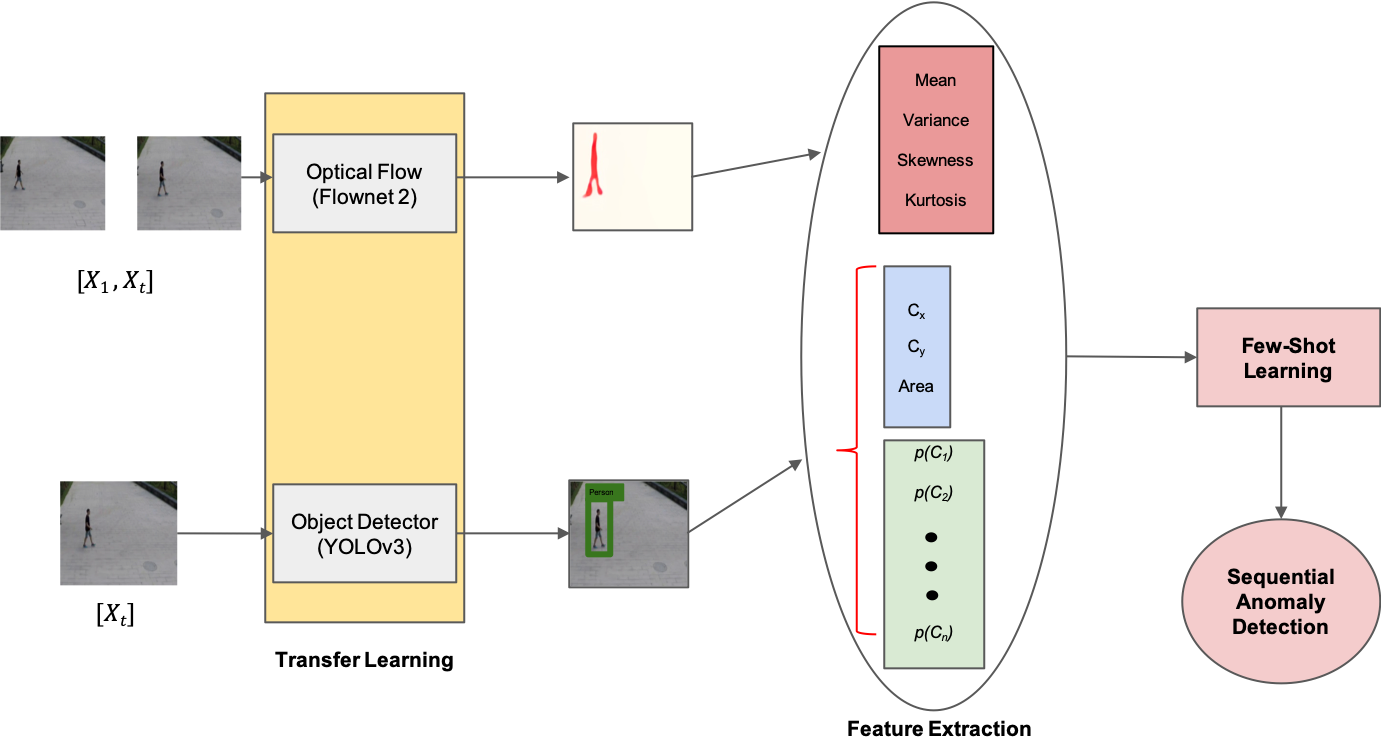}
\vspace{-2mm}
\caption{Proposed few-shot learning framework. At each time $t$, neural network-based feature extraction module provides motion (optical flow), location (center coordinates and area of bounding box), and appearance (class probabilities) features to the statistical anomaly detection module, which automatically sets its decision threshold to satisfy a false alarm constraint and makes online decisions.}
\label{f:system}
\vspace{-2mm}
\end{figure*}

An anomaly is construed as an unusual event which does not conform to the learned nominal patterns. However, in general, for practical implementations, it is unrealistic to assume the availability of sufficient training data for all possible nominal patterns/events. Thus, a practical framework should be able to perform \emph{any-shot} learning of nominal events. This presents a novel challenge to current approaches mentioned in Section \ref{s:Related} as their decision mechanism is extensively dependent on Deep Neural Networks (DNNs). DNNs typically require a large amount of training data with sufficient number of samples for each type of nominal event or exhibit the risk of catastrophic forgetting \cite{kirkpatrick2017overcoming}. Also, in general, the type of anomaly that the detector might encounter is broad and unknown while training the algorithm. For example, an anomalous event can be justified on the basis of appearance (a person carrying a gun), motion (two people fighting) or location (a person walking on the roadway). To account for all such cases, we create a feature vector $x^t_i$ for each object $i$ in frame $X^t$ at time $t$, where $x^t_i$ is given by $[w_1 x_{motion},w_2 x_{location},w_3 x_{appearance}]$. The weights $w_1, w_2, w_3$ are used to adjust the relative importance of each feature category. 

\subsection{Transfer Learning}
Most existing works propose training specialized data-hungry deep learning models from scratch, however this bounds their applicability to the cases where abundant data is available. Also, the training time required for such models grows exponentially with the size of training data, making them impractical to be deployed in scenarios where the model needs to continuously learn. Hence, we propose to leverage transfer learning to extract meaningful features from video. 
\label{s:detect}

\textbf{Object Detection:} To obtain location and appearance features, we propose to detect objects using a pre-trained real-time object detection system such as You Only Look Once (YOLO) \cite{redmon2016you}. YOLO offers a higher frames-per-second (fps) processing while providing better accuracy as compared to the other state-of-the-art models such as SSD and ResNet. For online anomaly detection, speed is a critical factor, and hence we currently prefer YOLOv3 in our implementations. For each detected object in image $X_t$, we get a bounding box (location) along with the class probabilities (appearance). Instead of simply using the entire bounding box, we monitor the center of the box and its area to obtain the location features. In a test video, objects diverging from the nominal paths and/or belonging to previously unseen classes will help us detect anomalies, as explained in Section \ref{s:anomaly}.

\textbf{Optical Flow:} Apart from spatial information, temporal information is also a critical aspect of videos. Hence, we propose to monitor the contextual motion of different objects in a frame using a pre-trained optical flow model such as Flownet 2 \cite{ilg2017flownet}. We hypothesize that any kind of motion anomaly would alter the probability distribution of the optical flow for the frame. Hence, we extract the mean, variance, and the higher order statistics skewness and kurtosis, which represent asymmetry and sharpness of the probability distribution.

Combining the motion, location, and appearance features, for each object detected in a frame, we construct a feature vector as shown in Fig. \ref{f:system}, where Mean, Variance, Skewness and Kurtosis are extracted from the optical flow; $C_x, C_y, Area$ denote the coordinates of the center of the bounding box and the area of the bounding box (Section \ref{s:detect}); and $p(C_1),\ldots,p(C_n)$ are the class probabilities for the detected object (Section \ref{s:detect}). Hence, at any given time $t$, with $n$ denoting the number of possible classes, the dimensionality of the feature vector is given by $D=n+7$. 

\subsection{Any-Shot Sequential Anomaly Detection}
\label{s:anomaly}

Anomaly detection in streaming video fits well to the sequential change detection framework \cite{basseville1993detection} as we can safely assume that any anomalous event would persist for an unknown but significant period of time. The eventual goal is to detect anomalies with minimal detection delays while satisfying a desired false alarm rate. Traditional parametric change detection algorithms which require probabilistic models cannot be implemented directly here as no prior knowledge about the anomalous events is available. Moreover, it is unrealistic to assume that the available training data includes sufficient number of frames for every possible nominal event. For example, while monitoring a street, the number of frames available for a car would be much more than a truck. 

\textbf{Training:} In the $N$-shot video setting, given a set of $N$ nominal frames, we leverage our transfer learning module to extract the training set %$\cS={(x_1,y_1),...,(x_M,y_M)}$, 
$\cS_M=\{x_1,...,x_M\}$, where $M$ is the number of detected objects, %from $C$ classes and a query set $Q={(\Tilde{x}_1,\Tilde{y}_1),...,(\Tilde{x}_{\Tilde{K}},\Tilde{y}_{\Tilde{K}})}$ where each $x_i,\tilde{x}_i \in\mathbb{R}^D$ 
and $x_i\in\mathbb{R}^D$ is a $D$-dimensional feature vector. %, and ($y_i,\tilde{y}_i$). 
Assuming that the training data does not include any anomalies, $\{x_1,\ldots,x_M\}$ correspond to $M$ points in the nominal data space, distributed according to an unknown complex probability distribution. To determine the nominal data patterns in a nonparametric way, we use $k$-nearest-neighbor ($k$NN) Euclidean distance to capture the interactions between the nominal data points due to its essential traits such as analyzability, interpretability, and computational efficiency, which deep learning-based models sorely lack. Given the informativeness of extracted motion, location, and appearance features, anomalous instances are expected to lie further away from the nominal training (support) set, which will lead to statistically higher $k$NN distances for the anomalous instances in the test (query) set with respect to the nominal data points. 
The training procedure of our detector is given as follows:

\begin{enumerate}
\item Partition the training set $\mathcal{S}_M$ into two sets $\mathcal{S}_{M_1}$ and $\mathcal{S}_{M_2}$ such that $M = M_1 + M_2$. 
\item Then for each feature vector $x_i$ in $\mathcal{S}_{M_1}$, we compute the $k$NN distance $d_i$ with respect to the points in $\mathcal{S}_{M_2}$. 
\item For a significance level $\alpha$, e.g., $0.05$, the $(1-\alpha)$th percentile $d_\alpha$ of $k$NN distances $\{d_1,\ldots,d_{M_1}\}$ is used as a baseline statistic for computing the anomaly evidence of test instances. 
\end{enumerate}

\textbf{Testing:} During the testing phase, for each object $i$ detected at time $t$, the sequential anomaly detection algorithm constructs the feature vector $x_i^t$ and computes the $k$NN distance $d_i^t$ with respect to the training instances in $\mathcal{S}$. Then, the instantaneous frame-level anomaly evidence $\delta^t$ is computed as
\begin{equation}
\label{eq:evidence}
    \delta^t = (\max_i\{d_i^t\})^D - d_\alpha^D.
\end{equation}
Finally, following a CUSUM-like procedure \cite{basseville1993detection} we update the running decision statistic $s_t$ as
\begin{equation}
    s^t = \max\{s^{t-1} + \delta^t,0\}, s^0 = 0.
\end{equation}

We decide that there exists an anomaly in video if the decision statistic $s^t$ exceeds the threshold $h$. After the anomaly decision, to determine the anomalous frames, we find the frame $s^t$ started to grow, say $\tau_{start}$, and also determine the frame $s_t$ stops increasing and keeps decreasing for a certain number, e.g., $5$, of consecutive frames, say $\tau_{end}$. Finally, we label the frames between $\tau_{start}$ and $\tau_{end}$ as anomalous, and continue testing for new anomalies with frame $\tau_{end}+1$ by resetting $s_{\tau_{end}}=0$.  

Existing works consider the decision threshold $h$ as a design parameter, however for a practical anomaly detection algorithm, a clear procedure for selecting it is necessary. In \cite{fap}, we provide an asymptotic ($M_2\to\infty$) upper bound on the false alarm rate:
\begin{equation}
    FAR \leq e^{-\omega_0h}, 
\end{equation}
where $\omega_0>0$ is given by
\begin{align}
    \label{e:thm}
    \omega_0 &= v_m - \theta -\frac{1}{\phi} \mathcal{W}\left( -\phi \theta e^{-\phi\theta } \right), \\
    \theta &= \frac{v_m}{e^{v_m d_\alpha^m}}.\nonumber
\end{align}
In \eqref{e:thm}, $\mathcal{W}(\cdot)$ is the Lambert-W function, $v_m=\frac{\pi^{m/2}}{\Gamma(m/2+1)}$ is the constant for the $m$-dimensional Lebesgue measure (i.e., $v_m d_\alpha^m$ is the $m$-dimensional volume of the hyperball with radius $d_\alpha$), and $\phi$ is the upper bound for $\delta_t$.
Although the expression for $\omega_0$ looks complicated, all the terms in \eqref{e:thm} can be easily computed. Particularly, $v_m$ is directly given by the dimensionality $m$, $d_\alpha$ comes from the training phase, $\phi$ is also found in training, and finally there is a built-in Lambert-W function in popular programming languages such as Python and Matlab. 
Hence, given the training data, $\omega_0$ can be easily computed, and the threshold $h$ can be chosen to asymptotically achieve the desired false alarm period as follows
\begin{equation}
h = \frac{-\log (FAR)}{\omega_0}.
\end{equation}

\section{Experiments}
\vspace{-1.5mm}
Most of the recent works evaluate their performance on three publicly available benchmark data sets, namely the UCSD pedestrian data set, the ShanghaiTech campus data set and the CUHK avenue data set. %The UCSD pedestrian dataset consists of 16 training and 12 test videos, each with a resolution of 240 x 360. The ShanghaiTech data set consists of 330 training and 107 test videos from 13 different scenes with a resolution of 480 x 85. The CUHK avenue data consists of 16 training and 21 test videos with a frame resolution of 360 x 640. 
Even though each data set has its own set of challenges, all the data sets have a common nominal definition. This makes them susceptible to trivial algorithmic designs which can achieve competitive results as there is a very obvious shift between the nominal and anomalous distributions. Hence, to make the problem more challenging and test the any-shot learning capabilities of different state-of-the-art algorithms, we also test on a modified version of the UCSD data set. For performance evaluation, following the existing works \cite{cong2011sparse,ionescu2019object,liu2018future}, we use the frame-level Area under the Curve (AuC) metric. %Next, we present our results on the \emph{benchmark} datasets and a unique \emph{any-shot learning} scenario.

\begin{figure}[th]
\centering
\includegraphics[width=0.48\textwidth]{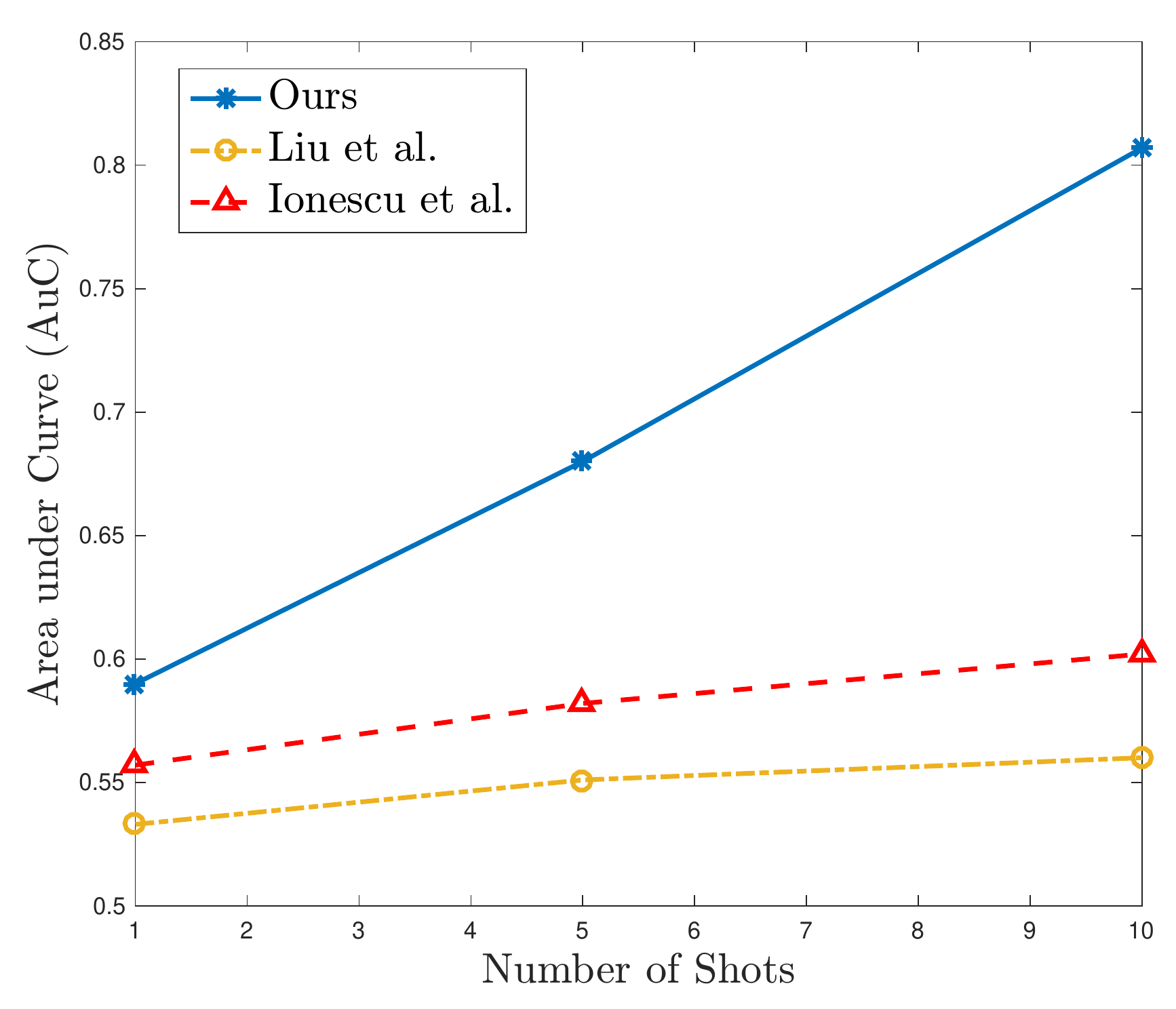}
\vspace{-3mm}
\caption{Comparison of the proposed and state-of-the-art algorithms Liu et al. \cite{liu2018future} and Ionescu et al. \cite{ionescu2019object} in terms of any-shot learning. The proposed algorithm is able to transition well between few-shot and many-shot learning.}
\label{f:any-shot}
%\vspace{-2mm}
\end{figure}

\textbf{Any-shot learning:} As compared to the original UCSD data set, where a person riding a bike is considered as anomalous, in this case we assume that it is a nominal behavior with very few training samples. Our goal here is to compare the any-shot learning capability of the proposed and state-of-the-art algorithms and see how well they adapt to new patterns. In this case, in addition to the available training data, we also train on a few samples of a person riding a bike. In Fig. \ref{f:any-shot}, it is seen that the proposed algorithm clearly outperforms the state-of-the-art algorithms \cite{ionescu2019object,liu2018future} in terms of any-shot learning performance. It is important to note that for video applications, 10 shots correspond to less than a second in real time.    

\vspace{-5.5mm}
\textbf{Benchmark Datasets:} To show the competitive performance of the proposed algorithm with large training data, we compare our results on the entire data sets to a wide range of methods in Table \ref{tab:my-table}. We should note here that our reported result in all the data sets is based on online decision making without seeing future video frames.
A common technique used by several recent works \cite{liu2018future,ionescu2019object} is to normalize the computed statistic for each test video independently, which is not suitable for online detection. %However, this methodology cannot be implemented in an online (real-time) system as it requires prior knowledge about the minimum and maximum values the statistic might take. 
%Recently, \cite{ionescu2019object} showed significant gains over the rest of the methods. However, their methodology of computing the AuC gives them an unfair advantage as they calculate the AuC for each video in a dataset, and then average them as the AuC of the dataset, as opposed to the other works which concatenate all the videos first and then determine the AuC as the data set's score.
\begin{table}[]
\centering
\resizebox{0.5\textwidth}{!}{%
\begin{tabular}{|c|c|c|c|c|}
\hline
Methodology    & CUHK Avenue & UCSD Ped 2 & ShanghaiTech  \\ \hline
Conv-AE \cite{hasan2016learning}        & 80.0        & 85.0       & 60.9\\ \hline
ConvLSTM-AE\cite{luo2017remembering}    & 77.0        & 88.1       & -   \\ \hline
Stacked RNN\cite{luo2017revisit}    & 81.7        & 92.2       & 68.0\\ \hline
GANs \cite{ravanbakhsh2018plug}           & -           & 88.4       & -   \\ \hline
Liu et al. \cite{liu2018future}     & 85.1        & 95.4       & 72.8\\ \hline
Sultani et al. \cite{sultani2018real} & -           & -          & 71.5\\ \hline
Ours           & 86.4        & 97.8       & 71.62     \\ \hline
\end{tabular}%
}
\caption{AuC result comparison on three datasets.}
\label{tab:my-table}
\vspace{-3mm}
\end{table}

% \subsection{Impact of Sequential Anomaly Detection}

% To demonstrate the importance of sequential anomaly detection in videos, we implement a nonsequential version of our algorithm by applying a threshold to the instantaneous anomaly evidence $\delta_t$, given in \eqref{eq:evidence}, which is similar to the approach employed by many recent works \cite{liu2018future,sultani2018real,ionescu2019object}. As Figure \ref{f:sequential} shows, instantaneous anomaly evidence is more prone to false alarms than the sequential MONAD statistic since it only considers the noisy evidence available at the current time to decide. Whereas, the proposed sequential statistic handles noisy evidence by integrating recent evidence over time. 

% \begin{figure}[th]
% \centering
% \includegraphics[width=0.5\textwidth]{Images/sequential.jpg}
% \vspace{-2mm}
% \caption{The advantage of sequential anomaly detection over single-shot detection in terms of controlling false alarms.}
% \label{f:sequential}
% \vspace{-2mm}
% \end{figure}

\vspace{-3mm}
\section{Conclusion}

For video anomaly detection, we presented an online anomaly detection algorithm which consists of a transfer learning-based feature extraction module and a statistical decision making module. The first module efficiently minimizes the training complexity and extracts motion, location, and appearance features. The second module is a sequential anomaly detector which enables a clear procedure for selecting decision threshold through asymptotic performance analysis. Through experiments on publicly available data, we showed that the proposed detector significantly outperforms the state-of-the-art algorithms in terms of any-shot learning of new nominal patterns. 0

{\small
\bibliographystyle{ieee_fullname.bst}
\bibliography{Ref.bib}

\begin{thebibliography}{10}\itemsep=-1pt

\bibitem{basseville1993detection}
Mich{\`e}le Basseville, Igor~V Nikiforov, et~al.
\newblock {\em Detection of abrupt changes: theory and application}, volume
  104.
\newblock prentice Hall Englewood Cliffs, 1993.

\bibitem{chandola2009anomaly}
Varun Chandola, Arindam Banerjee, and Vipin Kumar.
\newblock Anomaly detection: A survey.
\newblock {\em ACM computing surveys (CSUR)}, 41(3):1--58, 2009.

\bibitem{chaudhry2009histograms}
Rizwan Chaudhry, Avinash Ravichandran, Gregory Hager, and Ren{\'e} Vidal.
\newblock Histograms of oriented optical flow and binet-cauchy kernels on
  nonlinear dynamical systems for the recognition of human actions.
\newblock In {\em 2009 IEEE Conference on Computer Vision and Pattern
  Recognition}, pages 1932--1939. IEEE, 2009.

\bibitem{colque2016histograms}
Rensso Victor Hugo~Mora Colque, Carlos Caetano, Matheus Toledo~Lustosa de
  Andrade, and William~Robson Schwartz.
\newblock Histograms of optical flow orientation and magnitude and entropy to
  detect anomalous events in videos.
\newblock {\em IEEE Transactions on Circuits and Systems for Video Technology},
  27(3):673--682, 2016.

\bibitem{cong2011sparse}
Yang Cong, Junsong Yuan, and Ji Liu.
\newblock Sparse reconstruction cost for abnormal event detection.
\newblock In {\em CVPR 2011}, pages 3449--3456. IEEE, 2011.

\bibitem{dalal2005histograms}
Navneet Dalal and Bill Triggs.
\newblock Histograms of oriented gradients for human detection.
\newblock In {\em 2005 IEEE computer society conference on computer vision and
  pattern recognition (CVPR'05)}, volume~1, pages 886--893. IEEE, 2005.

\bibitem{fap}
Keval Doshi and Yasin Yilmaz.
\newblock Asymptotic upper bound on false alarm rate.
\newblock {\em \url{http://sis.eng.usf.edu/fap_supplementary.pdf}}.

\bibitem{hasan2016learning}
Mahmudul Hasan, Jonghyun Choi, Jan Neumann, Amit~K Roy-Chowdhury, and Larry~S
  Davis.
\newblock Learning temporal regularity in video sequences.
\newblock In {\em Proceedings of the IEEE conference on computer vision and
  pattern recognition}, pages 733--742, 2016.

\bibitem{hinami2017joint}
Ryota Hinami, Tao Mei, and Shin'ichi Satoh.
\newblock Joint detection and recounting of abnormal events by learning deep
  generic knowledge.
\newblock In {\em Proceedings of the IEEE International Conference on Computer
  Vision}, pages 3619--3627, 2017.

\bibitem{ilg2017flownet}
Eddy Ilg, Nikolaus Mayer, Tonmoy Saikia, Margret Keuper, Alexey Dosovitskiy,
  and Thomas Brox.
\newblock Flownet 2.0: Evolution of optical flow estimation with deep networks.
\newblock In {\em Proceedings of the IEEE conference on computer vision and
  pattern recognition}, pages 2462--2470, 2017.

\bibitem{ionescu2019object}
Radu~Tudor Ionescu, Fahad~Shahbaz Khan, Mariana-Iuliana Georgescu, and Ling
  Shao.
\newblock Object-centric auto-encoders and dummy anomalies for abnormal event
  detection in video.
\newblock In {\em Proceedings of the IEEE Conference on Computer Vision and
  Pattern Recognition}, pages 7842--7851, 2019.

\bibitem{jiang2018trust}
Heinrich Jiang, Been Kim, Melody Guan, and Maya Gupta.
\newblock To trust or not to trust a classifier.
\newblock In {\em Advances in neural information processing systems}, pages
  5541--5552, 2018.

\bibitem{kirkpatrick2017overcoming}
James Kirkpatrick, Razvan Pascanu, Neil Rabinowitz, Joel Veness, Guillaume
  Desjardins, Andrei~A Rusu, Kieran Milan, John Quan, Tiago Ramalho, Agnieszka
  Grabska-Barwinska, et~al.
\newblock Overcoming catastrophic forgetting in neural networks.
\newblock {\em Proceedings of the national academy of sciences},
  114(13):3521--3526, 2017.

\bibitem{koch2015siamese}
Gregory Koch, Richard Zemel, and Ruslan Salakhutdinov.
\newblock Siamese neural networks for one-shot image recognition.
\newblock In {\em ICML deep learning workshop}, volume~2. Lille, 2015.

\bibitem{kornblith2019better}
Simon Kornblith, Jonathon Shlens, and Quoc~V Le.
\newblock Do better imagenet models transfer better?
\newblock In {\em Proceedings of the IEEE conference on computer vision and
  pattern recognition}, pages 2661--2671, 2019.

\bibitem{krizhevsky2012imagenet}
Alex Krizhevsky, Ilya Sutskever, and Geoffrey~E Hinton.
\newblock Imagenet classification with deep convolutional neural networks.
\newblock In {\em Advances in neural information processing systems}, pages
  1097--1105, 2012.

\bibitem{videosurveillance}
L. Lin and N. Purnell.
\newblock A world with a billion cameras watching you is just around the
  corner.
\newblock {\em The Wall Street Journal,
  https://www.wsj.com/articles/a-billion-surveillance-cameras-forecast-to-be-watching-within-two-years-11575565402},
  2019.

\bibitem{liu2018future}
Wen Liu, Weixin Luo, Dongze Lian, and Shenghua Gao.
\newblock Future frame prediction for anomaly detection--a new baseline.
\newblock In {\em Proceedings of the IEEE Conference on Computer Vision and
  Pattern Recognition}, pages 6536--6545, 2018.

\bibitem{luo2017remembering}
Weixin Luo, Wen Liu, and Shenghua Gao.
\newblock Remembering history with convolutional lstm for anomaly detection.
\newblock In {\em 2017 IEEE International Conference on Multimedia and Expo
  (ICME)}, pages 439--444. IEEE, 2017.

\bibitem{luo2017revisit}
Weixin Luo, Wen Liu, and Shenghua Gao.
\newblock A revisit of sparse coding based anomaly detection in stacked rnn
  framework.
\newblock In {\em Proceedings of the IEEE International Conference on Computer
  Vision}, pages 341--349, 2017.

\bibitem{mo2013adaptive}
Xuan Mo, Vishal Monga, Raja Bala, and Zhigang Fan.
\newblock Adaptive sparse representations for video anomaly detection.
\newblock {\em IEEE Transactions on Circuits and Systems for Video Technology},
  24(4):631--645, 2013.

\bibitem{pan2009survey}
Sinno~Jialin Pan and Qiang Yang.
\newblock A survey on transfer learning.
\newblock {\em IEEE Transactions on knowledge and data engineering},
  22(10):1345--1359, 2009.

\bibitem{papernot2018deep}
Nicolas Papernot and Patrick McDaniel.
\newblock Deep k-nearest neighbors: Towards confident, interpretable and robust
  deep learning.
\newblock {\em arXiv preprint arXiv:1803.04765}, 2018.

\bibitem{ravanbakhsh2018plug}
Mahdyar Ravanbakhsh, Moin Nabi, Hossein Mousavi, Enver Sangineto, and Nicu
  Sebe.
\newblock Plug-and-play cnn for crowd motion analysis: An application in
  abnormal event detection.
\newblock In {\em 2018 IEEE Winter Conference on Applications of Computer
  Vision (WACV)}, pages 1689--1698. IEEE, 2018.

\bibitem{ravanbakhsh2017abnormal}
Mahdyar Ravanbakhsh, Moin Nabi, Enver Sangineto, Lucio Marcenaro, Carlo
  Regazzoni, and Nicu Sebe.
\newblock Abnormal event detection in videos using generative adversarial nets.
\newblock In {\em 2017 IEEE International Conference on Image Processing
  (ICIP)}, pages 1577--1581. IEEE, 2017.

\bibitem{ravanbakhsh2019training}
Mahdyar Ravanbakhsh, Enver Sangineto, Moin Nabi, and Nicu Sebe.
\newblock Training adversarial discriminators for cross-channel abnormal event
  detection in crowds.
\newblock In {\em 2019 IEEE Winter Conference on Applications of Computer
  Vision (WACV)}, pages 1896--1904. IEEE, 2019.

\bibitem{redmon2016you}
Joseph Redmon, Santosh Divvala, Ross Girshick, and Ali Farhadi.
\newblock You only look once: Unified, real-time object detection.
\newblock In {\em Proceedings of the IEEE conference on computer vision and
  pattern recognition}, pages 779--788, 2016.

\bibitem{sabokrou2018adversarially}
Mohammad Sabokrou, Mohammad Khalooei, Mahmood Fathy, and Ehsan Adeli.
\newblock Adversarially learned one-class classifier for novelty detection.
\newblock In {\em Proceedings of the IEEE Conference on Computer Vision and
  Pattern Recognition}, pages 3379--3388, 2018.

\bibitem{saligrama2012video}
Venkatesh Saligrama and Zhu Chen.
\newblock Video anomaly detection based on local statistical aggregates.
\newblock In {\em 2012 IEEE Conference on Computer Vision and Pattern
  Recognition}, pages 2112--2119. IEEE, 2012.

\bibitem{sitawarin2019defending}
Chawin Sitawarin and David Wagner.
\newblock Defending against adversarial examples with k-nearest neighbor.
\newblock {\em arXiv preprint arXiv:1906.09525}, 2019.

\bibitem{snell2017prototypical}
Jake Snell, Kevin Swersky, and Richard Zemel.
\newblock Prototypical networks for few-shot learning.
\newblock In {\em Advances in neural information processing systems}, pages
  4077--4087, 2017.

\bibitem{sultani2018real}
Waqas Sultani, Chen Chen, and Mubarak Shah.
\newblock Real-world anomaly detection in surveillance videos.
\newblock In {\em Proceedings of the IEEE Conference on Computer Vision and
  Pattern Recognition}, pages 6479--6488, 2018.

\bibitem{sung2018learning}
Flood Sung, Yongxin Yang, Li Zhang, Tao Xiang, Philip~HS Torr, and Timothy~M
  Hospedales.
\newblock Learning to compare: Relation network for few-shot learning.
\newblock In {\em Proceedings of the IEEE Conference on Computer Vision and
  Pattern Recognition}, pages 1199--1208, 2018.

\bibitem{vinyals2016matching}
Oriol Vinyals, Charles Blundell, Timothy Lillicrap, Daan Wierstra, et~al.
\newblock Matching networks for one shot learning.
\newblock In {\em Advances in neural information processing systems}, pages
  3630--3638, 2016.

\bibitem{xu2015learning}
Dan Xu, Elisa Ricci, Yan Yan, Jingkuan Song, and Nicu Sebe.
\newblock Learning deep representations of appearance and motion for anomalous
  event detection.
\newblock {\em arXiv preprint arXiv:1510.01553}, 2015.

\bibitem{yosinski2014transferable}
Jason Yosinski, Jeff Clune, Yoshua Bengio, and Hod Lipson.
\newblock How transferable are features in deep neural networks?
\newblock In {\em Advances in neural information processing systems}, pages
  3320--3328, 2014.

\bibitem{zhao2011online}
Bin Zhao, Li Fei-Fei, and Eric~P Xing.
\newblock Online detection of unusual events in videos via dynamic sparse
  coding.
\newblock In {\em CVPR 2011}, pages 3313--3320. IEEE, 2011.

\end{thebibliography}
}
\end{document}